\documentclass[10pt,twocolumn,letterpaper]{article}

\usepackage[pagenumbers]{cvpr} 

\usepackage{algorithm}
\usepackage{algpseudocode}
\usepackage[dvipsnames]{xcolor}

\usepackage{amssymb}
\usepackage{pifont}
\newcommand{\cmark}{\ding{51}}%
\newcommand{\xmark}{\ding{55}}%
\usepackage{multirow}
\usepackage{bm}

\definecolor{cvprblue}{rgb}{0.21,0.49,0.74}
\usepackage[pagebackref,breaklinks,colorlinks,citecolor=cvprblue]{hyperref}

\def\confName{CVPR}
\def\confYear{2025}

\title{DEFEND: A Large-scale 1M Dataset and Foundation Model for  \\Tobacco Addiction Prevention}

\author{Naga VS Raviteja Chappa$^{1}$, Matthew Shepard$^{1}$, Connor McCurtain$^{1}$, Charlotte McCormick$^{2}$, \\ Page Daniel Dobbs$^{2}$, Khoa Luu$^{1}$\\
$^{1}$Dept. of EECS, University of Arkansas\\
$^{2}$Center for Public Health and Technology, University of Arkansas\\ 
\tt\small \{nchappa, mjs042, crm063, cem044, pdobbs, khoaluu\}@uark.edu
}
\begin{document}

\twocolumn[{
 \renewcommand\twocolumn[1][]{#1}%
 \maketitle
 \begin{center}
 \centering
 \captionsetup{type=figure}
 \includegraphics[width=\textwidth]{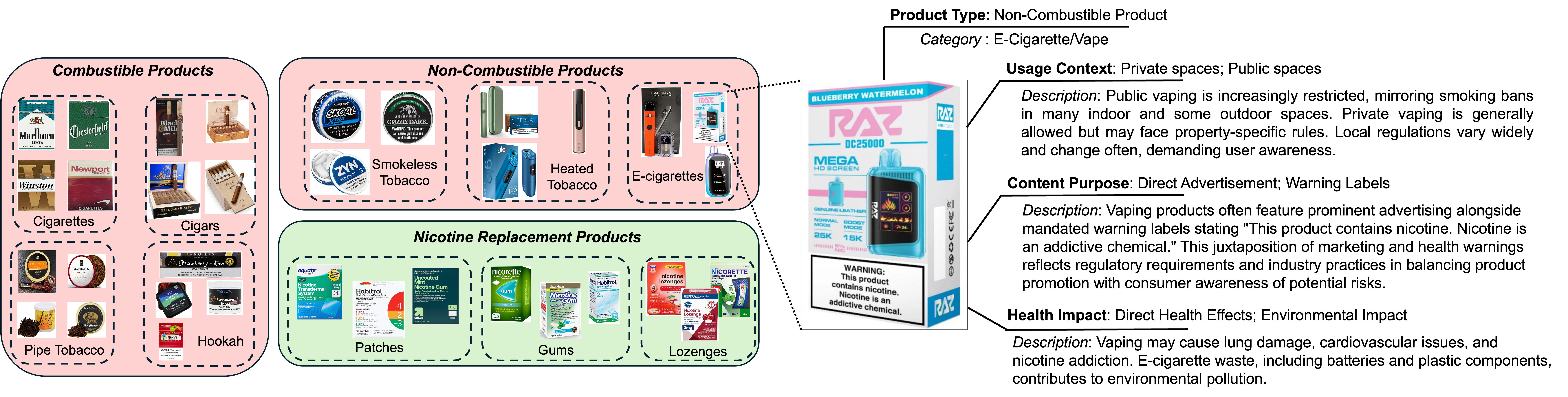}
 \captionof{figure}{{Examples of Our Tobacco-1M Dataset for Tobacco Product Understanding.} The left section displays visual samples from the major product categories, including \colorbox{red!20}{\textbf{Combustible Products}}, \colorbox{red!20}{\textbf{Non-Combustible Products}}, and \colorbox{green!20}{\textbf{Nicotine Replacement Products}}. The right section demonstrates a hierarchical annotation example of an E-cigarette product with detailed categorical descriptions. \textbf{Note:} This dataset is \emph{not intended} for the promotion of any tobacco products.
 }
 \label{fig:sample}
 \end{center}
 }]

\begin{abstract}
While tobacco advertising innovates at unprecedented speed, traditional surveillance methods remain frozen in time, especially in the context of social media. The lack of large-scale, comprehensive datasets and sophisticated monitoring systems has created a widening gap between industry advancement and public health oversight. This paper addresses this critical challenge by introducing Tobacco-1M, a comprehensive dataset of one million tobacco product images with hierarchical labels spanning 75 product categories, and DEFEND, a novel foundation model for tobacco product understanding. Our approach integrates a Feature Enhancement Module for rich multimodal representation learning, a Local-Global Visual Coherence mechanism for detailed feature discrimination, and an Enhanced Image-Text Alignment strategy for precise product characterization. Experimental results demonstrate DEFEND's superior performance, achieving 83.1\% accuracy in product classification and 73.8\% in visual question-answering tasks, outperforming existing methods by significant margins. Moreover, the model exhibits robust zero-shot learning capabilities with 45.6\% accuracy on novel product categories. This work provides regulatory bodies and public health researchers with powerful tools for monitoring emerging tobacco products and marketing strategies, potentially revolutionizing approaches to tobacco control and public health surveillance.
\end{abstract}

\section{Introduction}
Tobacco addiction persists as a critical public health challenge, with particularly severe implications for youth and young adults. The tobacco industry's rapid innovation in product development, including e-cigarettes and novel nicotine delivery systems, has consistently outpaced regulatory efforts and traditional monitoring methods. This dynamic landscape creates significant obstacles for public health officials, researchers, and policymakers in effectively addressing tobacco use and its far-reaching health impacts. Accurate detection and classification of tobacco products serve as a critical foundation for addiction prevention by enabling regulatory bodies to identify new products, enforce marketing restrictions, and develop targeted intervention strategies for vulnerable populations.

\begin{table*}[htbp]
\centering
\caption{Comparison with existing datasets related to tobacco products. Our proposed Tobacco-1M dataset has hierarchical labels with four main hierarchical levels, i.e., Product Type, Usage Context, Content Purpose, and Health Impact, and large numbers of product types and samples. Moreover, the proposed dataset contains hierarchical descriptions for each tobacco product and auxiliary categorization levels, i.e., Product Sub-categories, Direct health impacts, Marketing strategies, etc.}
\label{tab:tobacco-dataset-comparison}
\setlength{\tabcolsep}{9pt}
\small
\begin{tabular}{lcccccccc}
\hline
\textbf{Dataset} & \textbf{Year} & \begin{tabular}[c]{@{}c@{}}\textbf{Product}\\ \textbf{Types}\end{tabular} & \begin{tabular}[c]{@{}c@{}}\textbf{Hierarchical}\\ \textbf{Labels}\end{tabular} & \begin{tabular}[c]{@{}c@{}}\textbf{Hierarchical}\\ \textbf{Levels}\end{tabular} & \begin{tabular}[c]{@{}c@{}}\textbf{Product}\\ \textbf{Description}\end{tabular} & \begin{tabular}[c]{@{}c@{}}\textbf{Auxiliary}\\ \textbf{Categorization}\\ \textbf{Levels}\end{tabular} & \begin{tabular}[c]{@{}c@{}}\textbf{Number of}\\ \textbf{Samples}\end{tabular} \\
\hline
Murthy~\etal~\cite{ntad184} & 2024 & 3 & \textcolor{Red}{\xmark} & 1 & \textcolor{Red}{\xmark} & \textcolor{Red}{\xmark} & 826\\
Vassey~\etal~\cite{ntad224} & 2023 & 7 & \textcolor{Red}{\xmark} & 1 & \textcolor{Red}{\xmark} & \textcolor{Red}{\xmark} & 6,999\\
PHAD~\cite{Chappa_PHAD} & 2024 & 8 & \textcolor{Red}{\xmark} & 1 & \textcolor{Red}{\xmark} & \textcolor{Red}{\xmark} & 171,900\\
\textbf{Our Tobacco-1M} & \textbf{2024} & \textbf{75} & \textcolor{ForestGreen}{\cmark} & \textbf{4} & \textcolor{ForestGreen}{\cmark} & \textcolor{ForestGreen}{\cmark} & \textbf{1,128,652} \\
\hline
\end{tabular}%

\end{table*}

Existing tobacco product detection and classification methods have been constrained by limited datasets, often focusing on a narrow range of product types. Recent studies, such as Murthy et al.~\cite{ntad184}, Vassey et al.~\cite{ntad224}, and the PHAD dataset~\cite{Chappa_PHAD}, have made progress by expanding the number of product types examined. However, while valuable, these efforts still fall short of capturing the full spectrum of tobacco products in today's diverse market, thus limiting their utility in comprehensive tobacco control efforts.
 
The advent of foundation models in artificial intelligence offers a promising avenue for developing more robust and adaptable tobacco product monitoring systems. Recent advancements have demonstrated the power of these models, pre-trained on large-scale datasets, in revolutionizing vision tasks and generalizing well to various downstream applications~\cite{he2020momentum,chen2020improved,radford2021learning,jia2021scaling, caron2021emerging}. They excel at capturing both general and specific properties of visual data, making them potentially valuable tools for tobacco product analysis.

However, the development of effective foundation models for tobacco product monitoring faces a significant hurdle: the lack of a sufficiently large and diverse dataset. Current tobacco-related datasets, as evidenced by our comparative analysis (Table~\ref{tab:tobacco-dataset-comparison}), are limited in scale and scope. This limitation mirrors challenges observed in other domains, such as insect recognition~\cite{wu2019ip102}, where the diversity and complexity of the subject matter demand datasets of unprecedented scale and detail.

\noindent\textbf{Contributions of this Work:} 
To advance tobacco product monitoring and analysis, we introduce \textbf{Tobacco-1M}, a comprehensive tobacco product dataset, and \textbf{D}istillation-enabled \textbf{E}nhanced \textbf{F}eature learning for tobacco \textbf{EN}forcement and \textbf{D}iscernment (DEFEND), a novel foundation model for tobacco product understanding. First, we present Tobacco-1M, comprising one million images with hierarchical labels spanning from broad categories (e.g., Combustible, Non-Combustible) to specific types (e.g., Cigarettes, E-cigarettes), each annotated with detailed descriptions of features, usage context, and health impacts (refer to~\cref{fig:sample} for the dataset sample). This dataset is relatively 140 times larger than prior work~\cite{ntad224}. Second, we introduce a self-supervised learning approach with a Feature Enhancement module that captures nuanced correlations between visual and textual features. Third, we propose a Local-Global Visual Coherence loss that ensures consistency between fine-grained and holistic product representations. Fourth, we implement an Enhanced Image-Text Alignment mechanism with specialized contrastive loss for precise feature-description mapping. Finally, through extensive experiments on product classification, marketing strategy detection, and health impact assessment tasks, we demonstrate significant performance improvements over existing tobacco control and public health research methods.
\begin{figure*}[ht]
    \centering
    \includegraphics[width=\linewidth]{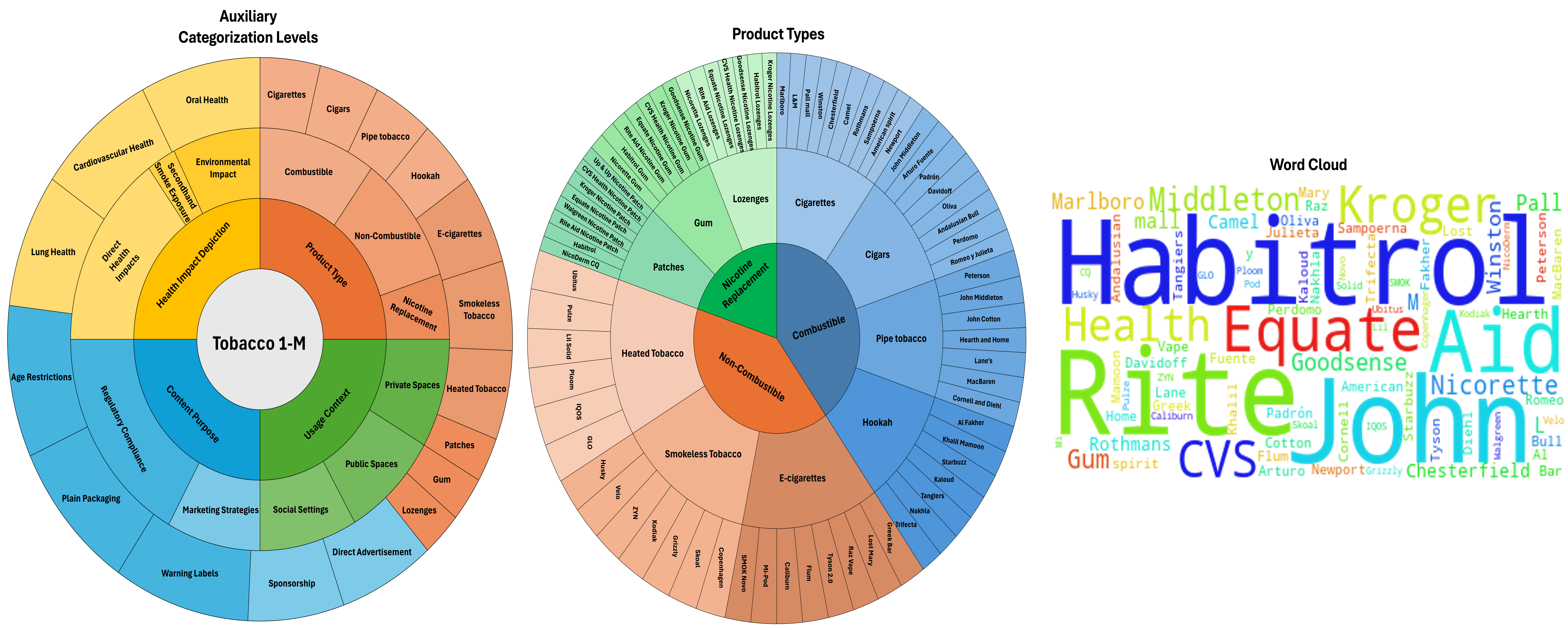}
    \caption{The Distribution of Tobacco Product Categories \emph{(left)} which illustrates the primary classification of our Tobacco-1M dataset with its four major domains, while The Distribution of Product Type Hierarchy \emph{(center)} provides a detailed breakdown of product types and their associated brands. Word Cloud \emph{(right)} of the most frequent product types in our dataset. \textbf{Best viewed with zoom-in and color.}}
    \label{fig:stats}
\end{figure*}

\section{Related Works}
Recent advancements in video analysis~\cite{wang2021actionclip, chappa2023sogar, chappa2023spartan, tang2018vctree, zhong2024learning, zareian2020bridging, zareian2020learning, lu2016visual, zhong2021learning, ye2021linguistic, nguyen2024type} have predominantly employed traditional deep learning algorithms, but the absence of language comprehension limits their efficacy. Recognizing this gap, there is a shift towards integrating text modalities~\cite{chappa2024react, chappa2024hatt} to enhance understanding capabilities. It underscores the need to evolve existing algorithms, advocating for joint training with Large Language Models (LLMs)~\cite{liu2024visual} to achieve superior performance and nuanced comprehension.

\noindent\textbf{Foundation Models and Multimodal Learning.}
Recent foundation model research~\cite{xu2023mplug, chen2024feddat, luo2023molfm} has advanced multimodal approaches, integrating diverse data sources for richer representation learning. These models have demonstrated particular success in hierarchical learning~\cite{mnih2008scalable, alper2024emergent}, enabling classification across both broad and fine-grained categories. Transfer learning capabilities~\cite{kang2023grounding, garau2024multi} have further extended their utility, allowing models trained on one domain to generalize to related tasks. For instance, Luis~\etal~\cite{garau2024multi} successfully applied transfer learning to connect training across DNA, RNA, and protein sequences, demonstrating the potential for cross-domain knowledge transfer.

\noindent\textbf{Tobacco Content Analysis in Social Media.}
Recent studies~\cite{kong2023understanding, murthy2023influence, chappa2024advanced} have employed machine learning techniques to analyze e-cigarette content and promotion strategies on platforms like YouTube, focusing on how user profiles influence content engagement. Prior work~\cite{ntad224} developed computer vision models using Instagram data (6,999 images) labeled for e-cigarette-related objects, while Murthy~\etal~\cite{ntad184} extended this to TikTok using YOLOv7~\cite{wang2023yolov7} for detecting vaping devices and related content in 826 annotated images.

\noindent\textbf{AI Applications in Tobacco Research.}
Machine learning has been increasingly applied to broader tobacco research domains. Studies have utilized regression tree models for analyzing global tobacco survey data~\cite{kim2021machine}, identifying factors contributing to adolescent tobacco use. Other research has focused on smoking cessation~\cite{perski2023classification}, using machine learning to predict and analyze relapse patterns through app-based data collection. Recent work~\cite{lakatos2024multimodal} has demonstrated the potential of deep learning in detecting covert tobacco advertisements through integrated multimodal analysis of text and images.

\noindent\textbf{Large-scale Dataset Development.}
While existing datasets have contributed to tobacco content analysis, they remain limited in scale and scope. Current public datasets range from 826 images~\cite{ntad184} to 6,999 images~\cite{ntad224}, focusing primarily on specific product categories or platforms. This limitation mirrors challenges observed in other domains~\cite{wu2019ip102, nguyen2024insect}, where the complexity of the subject matter demands larger, more comprehensive datasets for effective model training.
Our work builds upon these foundations while addressing their limitations by introducing Tobacco-1M, a large-scale dataset comprising 1 million images across 75 product categories, and DEFEND, a novel foundation model designed explicitly for understanding tobacco products.

\section{The Proposed Tobacco 1M Dataset}

The Tobacco-1M dataset represents a significant advance in the field of tobacco product analysis and addiction control. This large-scale and diverse dataset has been meticulously curated to address the limitations of existing datasets, as shown in~\Cref{tab:tobacco-dataset-comparison}, and to provide a comprehensive resource for researchers and policymakers in the domain of tobacco control.

\subsection{Dataset Overview}

Tobacco-1M comprises more than one million high-resolution images of tobacco products, encompassing a wide range of categories and subcategories, as shown in~\cref{fig:stats} (right). The data set is structured to allow a nuanced analysis of various aspects of tobacco products and their impacts on usage. For a comprehensive description of the dataset's structure and format, please refer to Sec. \textcolor{red}{B} of the Supplementary Material (Supp.).

\subsection{Data Collection and Preprocessing}

The images in Tobacco-1M are collected through a rigorous process involving systematic sampling of tobacco products from various online sources and official product catalogs (detailed in Sec. \textcolor{red}{A} of Supp.). Each sample in the dataset is assigned a tobacco product image and its corresponding taxonomic label. Here, taxonomic labels are hierarchical in structure as demonstrated in~\cref{fig:sample}. Our team of public health experts \textit{manually verified and annotated} each entry to ensure the accuracy of the data.

\subsection{Dataset Statistics}
\Cref{fig:stats} (left) illustrates the distribution of auxiliary categorization levels in Tobacco-1M, with Product Type dominating. It also shows subcategory distributions, notably a balanced representation of health concerns within the Health Impact Depiction category. \Cref{fig:stats} (center) breaks down Product Types and brands, revealing combustible products, especially cigarettes, as the majority. Brand distribution within each Product Type is comprehensive. The dataset balances real-world product prevalence with diversity, enabling robust AI model development across the tobacco product spectrum.

\subsection{Significance and Potential Applications}
Tobacco-1M's comprehensive categorization and diverse coverage facilitate the development of generalizable AI models for tobacco product detection, classification, and trend analysis. It supports research on product evolution, health impacts, and regulatory compliance across jurisdictions (refer to Sec. \textcolor{red}{G} of Supp.). This dataset aims to accelerate AI-driven tobacco monitoring research, contributing to effective control policies. Tobacco-1M serves as a vital resource for researchers, policymakers, and health professionals in reducing tobacco-related disease burden globally. 

\section{Methodology}

\subsection{Limitations in Prior Training Approaches}

A critical challenge in tobacco product understanding is the accurate representation and interpretation of subtle product features that distinguish different categories and variants. While CLIP \cite{radford2021learning} demonstrates strong performance in general visual-language tasks, it struggles to capture fine-grained product attributes and regulatory elements crucial for tobacco product analysis. Similarly, vision-language models like CoCa \cite{yu2022coca} process images holistically, often missing the detailed textural and structural characteristics that differentiate tobacco products. 

\begin{figure}[ht]
    \centering
    \includegraphics[width=\linewidth]{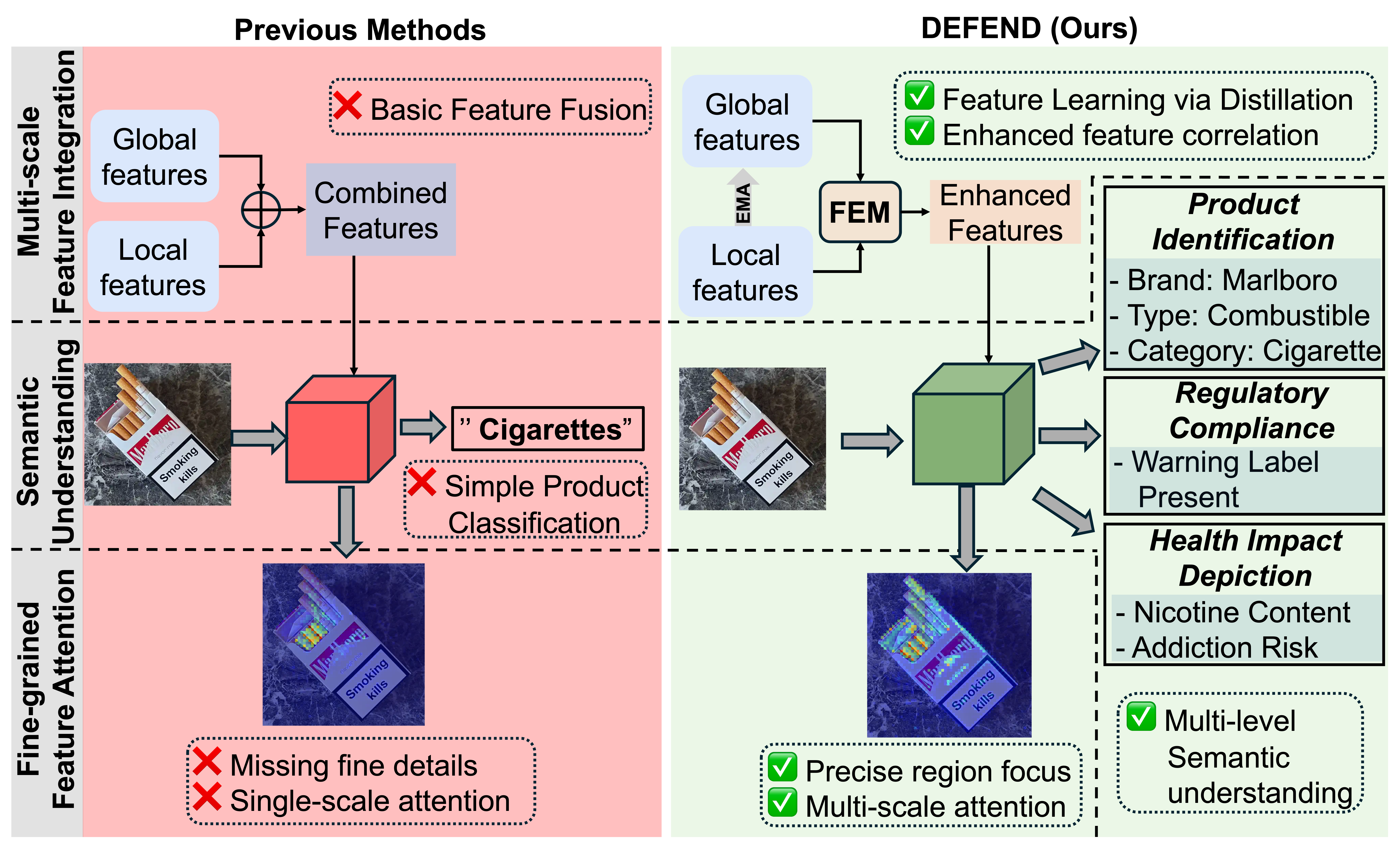}
    \caption{\textbf{Comparisons of Previous Methods.} Prior works~\cite{yu2022coca, he2022masked, fan2021multiscale, 9779951} demonstrate significant limitations in tobacco product analysis: naive feature fusion fails to capture nuanced product characteristics due to oversimplified integration, basic classification architecture overlooks critical regulatory elements. DEFEND overcomes these limitations through Feature Enhancement Module (FEM), enabling precise feature correlation across scales, that distinguishes critical regulatory elements and health impact indicators through targeted region focus and multi-scale attention mechanisms. \textbf{Best viewed with zoom in and color.}}
    \label{fig:limitations}
\end{figure}

Current approaches using masked image modeling \cite{he2022masked} focus primarily on broad feature reconstruction without explicitly addressing the nuanced visual elements of tobacco products, which is visualized in~\cref{fig:limitations}. While hierarchical vision transformers \cite{fan2021multiscale} attempt to capture multi-scale features, they lack mechanisms to combine local product details with global contextual understanding effectively. Furthermore, existing contrastive learning methods \cite{9779951} emphasize general semantic alignment between images and text, but fail to capture the specialized relationships between tobacco product attributes and their technical descriptions.

\begin{figure*}[ht]
    \centering
    \includegraphics[width=0.9\linewidth]{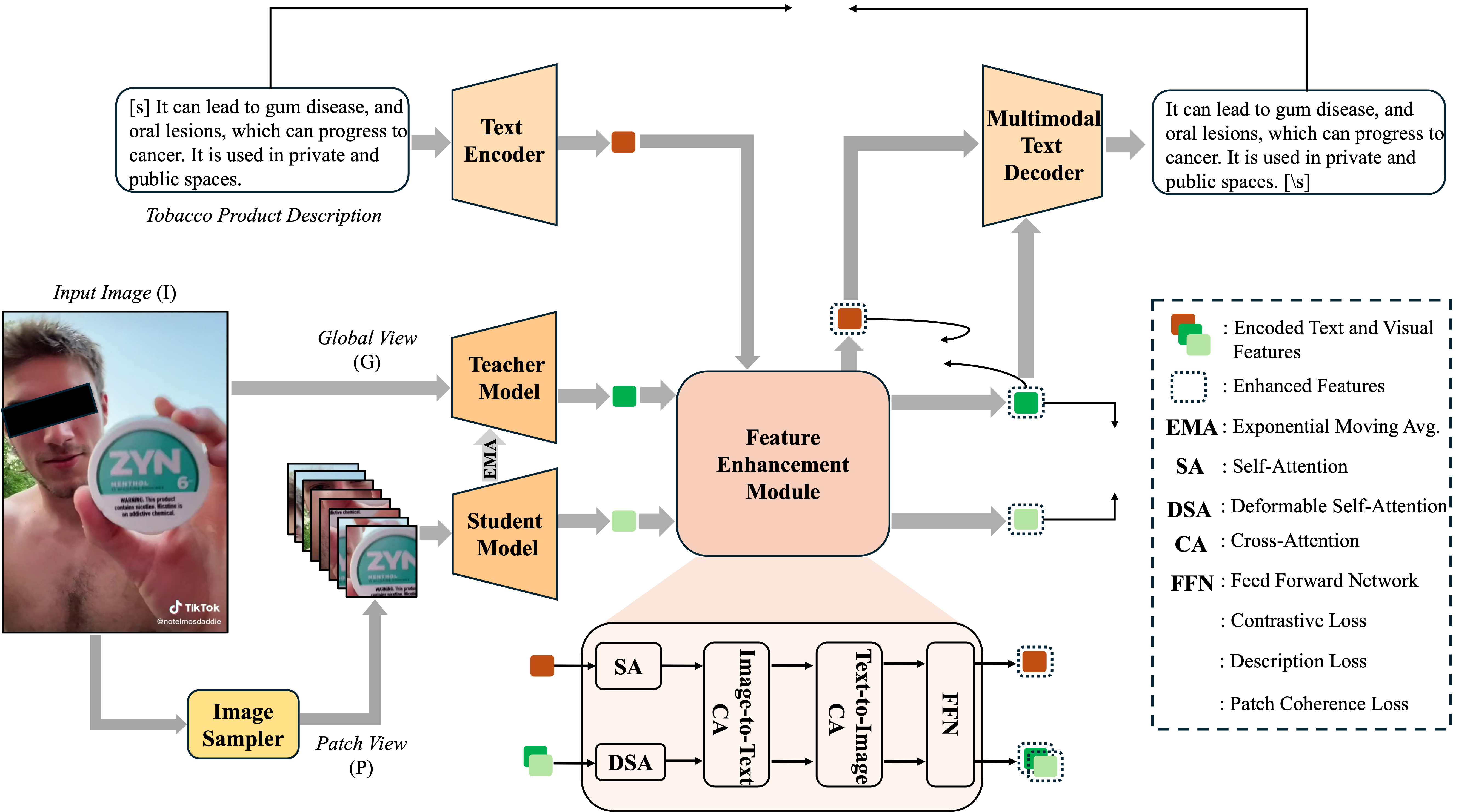}
    \put(-263,133){\small$\bm{f_{G}}$}
    \put(-263,95){\small$\bm{f_{P}}$}
    \put(-263,212){\small$\bm{f_{T}}$}
    \put(-206,158){\small$\bm{f_{T}^*}$}
    \put(-129,129){\small$\bm{f_{G}^*}$}
    \put(-129,80){\small$\bm{f_{P}^*}$}
    \put(-245,48){\scriptsize{K,V}}
    \put(-243,8){\scriptsize{Q}}
    \put(-210,48){\scriptsize{Q}}
    \put(-213,8){\scriptsize{K,V}}
    \put(-117,106){\small$\bm{\mathcal{L}_{pc}}$}
    \put(-180,139){\small$\bm{\mathcal{L}_{con}}$}
    \put(-206,239){\small$\bm{\mathcal{L}_{desc}}$}
    \put(-96, 56){\scriptsize$\bm{\mathcal{L}_{con}}$}
    \put(-96,41){\scriptsize$\bm{\mathcal{L}_{desc}}$}
    \put(-95,30){\scriptsize$\bm{\mathcal{L}_{pc}}$}
    \caption{\textbf{Overview of our proposed DEFEND Framework.} We employ a dual-stream design where text input is processed through a Text Encoder while image input follows two parallel paths through a Teacher-Student architecture. The Feature Enhancement Module enhances these multimodal features, which are then used to train the model with multiple objectives, including contrastive, description, and patch coherence losses, to generate comprehensive multimodal representations.}
    \label{fig:framework}
\end{figure*}
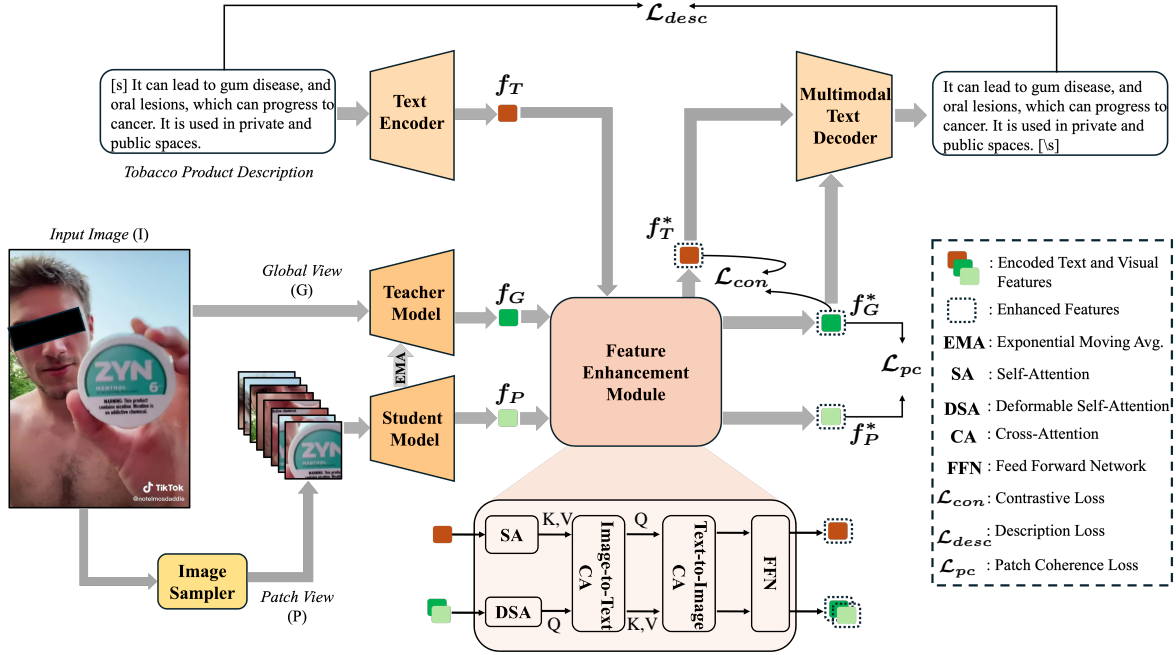

To overcome these challenges, we introduce DEFEND (as illustrated in~\cref{fig:framework}), a novel framework that synergizes global product understanding with local feature discrimination through an innovative teacher-student architecture. Our approach not only enhances the model's ability to capture subtle product characteristics but also maintains semantic consistency across different scales of visual analysis. 

\subsection{Input Modeling}


Given an input image $I \in \mathbb{R}^{H \times W \times 3}$, we process it into two complementary representations, including a global view $G$ and a set of local patches $P = \{p_i\}_{i=1}^{N_P}$. The number of patches is determined by $N_P = HW / s_p^2$, where $s_p \times s_p$ is the patch resolution. To enhance feature discrimination, we employ an adaptive sampling strategy as in
\(
    P_s = \Psi(P, \lambda) \subset P
\), %
where $\Psi$ is our adaptive sampling function (detailed in Sec. \textcolor{red}{C.1} of Supp.) that selects patches based on saliency parameter $\lambda$, focusing on regions with distinctive product characteristics. This dual representation enables our model to simultaneously capture global context and local details crucial for tobacco product analysis.
\subsection{Teacher-Student Architecture}

Our framework employs a teacher-student architecture where knowledge is distilled from the student to the teacher through Exponential Moving Average (EMA) updates while maintaining a stable global view of tobacco products. This design helps capture broad product categories and subtle distinguishing features like warning labels and brand elements.

\subsubsection{Teacher Model - The Global Encoder}
The teacher model $E_{global}$ serves as a stable global feature extractor. It processes the entire image to establish a holistic understanding, as in Eqn. \eqref{eqn:glo}.
\begin{equation} \label{eqn:glo}
    \bm{f_G} = E_{global}(G), \quad \bm{f_G} \in \mathbb{R}^{1 \times D}
\end{equation}
where $D$ is the feature dimension. Knowledge is distilled from the student to the teacher through EMA updates of the network parameters: \(\theta_t^{(k)} = \alpha\theta_t^{(k-1)} + (1-\alpha)\theta_s^{(k)}\)
where $\theta_t^{(k)}$ and $\theta_s^{(k)}$ are the teacher and student parameters at iteration $k$ respectively, and $\alpha \in [0,1]$ is the EMA decay rate (typically set to 0.999). 

\subsubsection{Student Model - The Patch Encoder}
The student model $E_{patch}$ learns to extract discriminative local features under the teacher's guidance. For each sampled patch $p_i \in P_s$, it generates features encouraged to align with the teacher's global understanding, as in Eqn. \eqref{eqn:Xs}.
\begin{equation} \label{eqn:Xs}
    \bm{X_s} = \text{concat}[\bm{x_i}]_{i=1}^{N_{P_s}} \in \mathbb{R}^{N_{P_s} \times D}, \quad \bm{x_i} = \alpha_p(p_i) + \bm{e_p}(i)
\end{equation}
where $\alpha_p$ is a patch projection function and $\bm{e_p}$ is a learnable position embedding that maintains spatial relationships between patches. The student processes these patches through $L_e$ transformer blocks as follows,
\begin{align} \label{eqn:Le}
    \bm{X}_l' &= \bm{X}_{l-1} + \text{MSA}(\text{LN}(\bm{X}_{l-1})) \\
    \bm{X}_l &= \bm{X}_l' + \text{FFN}(\text{LN}(\bm{X}_l')) \\
    \bm{X}_0 &= \bm{X}_s, \quad 1 \leq l \leq L_e
\end{align}
where MSA denotes multi-head self-attention, LN is layer normalization, and FFN represents a feed-forward network. The final patch-level features are obtained as in Eqn. \eqref{eqn:fP}.
\begin{equation} \label{eqn:fP}
    \bm{f_P} = E_{patch}(\bm{X}_s), \quad \bm{f_P} \in \mathbb{R}^{N_{P_s} \times D}
\end{equation}
Through this EMA-based knowledge transfer, the student learns to extract local features while the teacher maintains a stable global view. 

\subsection{Text Encoder}

The text encoder processes input descriptions to create representations that can effectively align with our visual features. Given a text description $T = \{t_i\}_{i=1}^{N_T}$, where $N_T$ is the number of tokens, we employ BERT~\cite{devlin2018bert} as our text encoder $E_{\text{text}}$ to generate contextual representations as follows,
\begin{equation} \label{eqn:fT}
    \bm{f_T} = E_{\text{text}}(\bm{T}), \quad \bm{f_T} \in \mathbb{R}^{N_T \times D}
\end{equation}
%
%
\begin{align}
    \bm{H}_l &= \text{MSA}(\text{LN}(\bm{f_T}^{l-1})) + \bm{f_T}^{l-1} \\
    \bm{f_T}^l &= \text{FFN}(\text{LN}(\bm{H}_l)) + \bm{H}_l, \quad l = 1, ..., L_t
\end{align}
where $L_t$ is the number of transformer layers, MSA denotes multi-head self-attention, LN is layer normalization, FFN is a feed-forward network, and $f_T^0$ represents the initial token embeddings. 

\subsection{Feature Enhancement Module}
The Feature Enhancement Module (FEM) enriches both visual and textual representations through a unified multi-stage attention mechanism. Given global features $\bm{f_G} \in \mathbb{R}^{1 \times D}$ from the teacher and patch features $\bm{f_P} \in \mathbb{R}^{N_{P_s} \times D}$ from the student, we process these representations through intra-modal and cross-modal attention stages.

The feature enhancement process begins with modality-specific self-attention as follows,
\begin{align}
    \hat{f}_T &= A_T(\bm{f_T}, \bm{f_T}, \bm{f_T}), \\ 
    \hat{f}_V &= A_V(\bm{f_V}, \bm{f_V}, \bm{f_V}), \quad V \in \{G,P\}
\end{align}
where $A_T$ and $A_V$ are modality-specific self-attention mechanisms. The cross-modal attention and refinement then proceed as follows,
\begin{align}
    \bm{f_T^*} &= F_T(A_{TV}(\hat{f}_T, \hat{f}_V, \hat{f}_V)) \\
    \bm{f_V^*} &= F_V(A_{VT}(\hat{f}_V, \hat{f}_T, \hat{f}_T)), \quad V \in \{G,P\}
\end{align}
where $A_{TV}$ and $A_{VT}$ are cross-modal attention operations, and $F_T$ and $F_V$ are feed-forward networks for final feature refinement. Each attention mechanism follows the scaled dot-product attention: $\text{Attention}(\bm{Q},\bm{K},\bm{V}) = \text{softmax}(\frac{\bm{Q}\bm{K}^T}{\sqrt{d}})\bm{V}$, where $\bm{Q}$, $\bm{K}$, and $\bm{V}$ represent query, key, and value matrices respectively, and $d$ is the feature dimension.
\subsection{Enhanced Image-Text Alignment}

To further improve the alignment between visual and textual representations, inspired by recent advances in vision-language modeling~\cite{radford2021learning,yu2022coca}, we employ a contrastive learning strategy. While these works demonstrate the effectiveness of contrastive learning for general image-text alignment, we adapt this approach specifically for tobacco product understanding by leveraging our enriched multimodal features from FEM.

The contrastive loss $\mathcal{L}_{cont}$ encourages semantic alignment between matched image-text pairs while distinguishing unrelated pairs as
\(    \mathcal{L}_{cont} = -\log \frac{\exp(s(\bm{f_T^*}, \bm{f_G^*}) / \tau)}{\sum_{k} \exp(s(\bm{f_T^*}, \bm{f_G^*}{_k}) / \tau)}
\),
where $s(\cdot, \cdot)$ denotes cosine similarity, $\tau$ is a temperature parameter, and $\bm{f_G^*}{_k}$ represents global features from other samples in the batch. By applying this loss to our enhanced features $\bm{f_T^*}$ and $\bm{f_G^*}$, we leverage the rich representations learned through our FEM to create more precise alignments between product images and their technical descriptions.

\subsection{Local-Global Visual Coherence}

Maintaining consistency between fine-grained details and overall product characteristics in tobacco product analysis is essential for accurate classification and attribute detection. Local features must align with the global context while preserving discriminative details specific to product categories and variants. To achieve this balance, we introduce a patch coherence mechanism.

The patch coherence loss $\mathcal{L}_{pc}$ ensures alignment between local and global representations as
%
\(    \mathcal{L}_{pc} = \| \bm{f_G^*} - \text{AvgPool}(\bm{f_P^*}) \|_2^2\).
%
This loss term ensures that local patch representations contribute coherently to the global product understanding while maintaining their distinctive characteristics.

\subsection{Multimodal Image Description Decoder}

While contrastive learning establishes broad semantic alignments, but precise description of tobacco products requires capturing intricate product attributes, regulatory elements, and interrelationships. Our decoder architecture addresses this by incorporating enhanced visual and textual representations in a structured generation process.

The multimodal description generation process and training objective are formalized as follows:
\begin{align}
    h_j &= \text{DecoderLayer}([y_{<j}; \bm{f_T^*}; \bm{f_G^*}]) \\
    p(y_j | y_{<j}) &= \text{softmax}(\bm{W}_o h_j + b_o) \\
    \mathcal{L}_{desc} &= - \frac{1}{N}\sum_{i=1}^{N} \sum_{j=1}^{M_i} \log p(y_{ij} | y_{i,<j}, \bm{f_T^*}, \bm{f_G^*})
\end{align}
%
where $h_j$ is the decoder hidden state, $\bm{W_o}$ and $b_o$ are learnable parameters, $N$ is the batch size, and $M_i$ is the length of the $i$-th description. Please refer to Sec. \textcolor{red}{B} of Supp. for a detailed Inference procedure.

\section{Experimental Results}

\subsection{Implementation Details}

\noindent\textbf{Datasets.} We conduct our main experiments on the PHAD dataset~\cite{Chappa_PHAD}, which contains 5,730 videos sampled at 24 frames per second following their experiments across eight categories. For pre-training, we utilize our newly collected Tobacco-1M dataset, comprising 1M high-resolution tobacco product images across 75 categories: combustible products (450K images), non-combustible products (200K images), and nicotine replacement products (50K images). Each image is annotated with comprehensive metadata, including product type, brand information, and health warnings. We use 700K images for training, reserving 150K for VQA task evaluation and 150K as a testing set.

\noindent\textbf{Model Configurations.}
Our framework adopts ViT-B/16~\cite{dosovitskiy2020image} as the backbone, processing 224×224 images with a 16×16 patch size. The model integrates an adaptive sampling mechanism ($\lambda=0.3$) that prioritizes regions with product-distinctive features (e.g., warning labels, brand logos). For text processing, we leverage BERT~\cite{devlin2018bert} as our encoder and extend it with cross-attention layers for visual-textual fusion. Training proceeds in three phases on 4×A100 GPUs: warm-up (20 epochs), main training (200 epochs), and fine-tuning (80 epochs), using AdamW~\cite{loshchilov2017decoupled} with 1e-3 learning rate and cosine decay~\cite{loshchilov2016sgdr}. Please refer to Sec. \textcolor{red}{C} of the Supp. for comprehensive implementation details, including hyperparameter selection, data augmentation strategies, and training procedures.

\noindent\textbf{Metrics.} We evaluated our model using standard classification metrics: accuracy, precision, recall, and F1-score. For the visual question answering (VQA) task, we used accuracy and F1-score. Zero-shot performance was assessed using top-1 and top-5 accuracy on unseen classes.

\subsection{Ablation Studies}
Our comprehensive ablation study, presented in Table~\ref{tab:ablation}, reveals the individual and combined effects of various components in our Tobacco Foundation Model on the Tobacco-1M Classification task. Please refer to Sec. \textcolor{red}{D} of the Supp. for more ablation studies.

\begin{table}[h]
\centering
\caption{Ablation study results on the Tobacco-1M dataset. Here, Acc.: Accuracy, FEM: Feature Enhancement Module, $\mathcal{L}_\text{pc}$: Patch Coherence Loss, $\mathcal{L}_\text{con}$: Contrastive Loss, and $\mathcal{L}_\text{desc}$: Description Loss.}
\label{tab:ablation}
\setlength{\tabcolsep}{4pt}
\small
\begin{tabular}{lccccccc}
\hline
\textbf{Backbone} & \textbf{FEM} & \textbf{$\mathcal{L}_\text{pc}$} & \textbf{$\mathcal{L}_\text{con}$} & \textbf{$\mathcal{L}_\text{desc}$} & \multirow{2}{*}{\begin{tabular}[c]{@{}c@{}}\textbf{Acc.@1}\\\textbf{(\%)}\end{tabular}} & \multirow{2}{*}{\begin{tabular}[c]{@{}c@{}}\textbf{Acc.@5}\\\textbf{(\%)}\end{tabular}} \\
 & & & & & & \\
\hline
& $\checkmark$ & & & & 71.2 & 76.5 \\
ViT-small/16 & $\checkmark$ & $\checkmark$ & & & 72.4 & 77.3 \\
& $\checkmark$ & $\checkmark$ & $\checkmark$ & & 73.5 & 78.2 \\
& $\checkmark$ & $\checkmark$ & $\checkmark$ & $\checkmark$ & 74.2 & 79.1 \\
\hline
& $\checkmark$ & & & & 73.8 & 78.6 \\
ViT-base/16 & $\checkmark$ & $\checkmark$ & & & 75.1 & 79.8 \\
& $\checkmark$ & $\checkmark$ & $\checkmark$ & & 77.1 & 81.9 \\
& $\checkmark$ & $\checkmark$ & $\checkmark$ & $\checkmark$ & \textbf{78.3} & \textbf{83.1} \\
\hline
& $\checkmark$ & & & & 73.2 & 78.1 \\
ViT-large/16 & $\checkmark$ & $\checkmark$ & & & 74.5 & 79.3 \\
& $\checkmark$ & $\checkmark$ & $\checkmark$ & & 76.2 & 81.1 \\
& $\checkmark$ & $\checkmark$ & $\checkmark$ & $\checkmark$ & 77.4 & 82.3 \\
\hline
\end{tabular}
\small{\textbf{Note:} \emph{FEM Only} experiments uses basic cross-entropy loss.}
\end{table}
\noindent\textbf{Effectiveness of Backbone Networks.}
The study encompasses three Vision Transformer variants: ViT-small/16, ViT-base/16, and ViT-large/16. Our findings indicate that the ViT-base/16 architecture achieves the best performance, with a peak Top-1 accuracy of 78.3\%. This represents a substantial improvement over both the ViT-small/16 (74.2\%) and ViT-large/16 (77.4\%) models, suggesting that this architecture provides an optimal balance between model capacity and efficiency in capturing the nuanced features of tobacco products.

\noindent\textbf{Effectiveness of FEM.}
The Feature Enhancement Module (FEM) emerges as a cornerstone of our model's success. Its integration yields significant performance boosts across all architectures, establishing strong baseline accuracies of 71.2\%, 73.8\%, and 73.2\% for small, base, and large models respectively. These results highlight the crucial role of FEM in extracting and refining relevant features from tobacco product images with the help of textual descriptions.

\noindent\textbf{Effectiveness of Enhanced Image-Text
Contrastive Loss.}
The incorporation of image-text contrastive loss proves highly beneficial, driving accuracy improvements of 1.1\%, 2.0\%, and 1.7\% for small, base, and large models, respectively. This consistent performance gain across architectures demonstrates the effectiveness of aligning visual and textual representations, enabling the model to better differentiate between visually similar tobacco products.

\noindent\textbf{Effectiveness of Description Loss.}
Our analysis reveals that the Description Loss component further refines the model's capabilities. It yields additional accuracy improvements of 0.7\%, 1.2\%, and 1.2\% across the three model sizes. These gains suggest enhanced alignment between visual features and product descriptions, contributing to more precise classification, particularly for products with subtle distinctions.

\noindent\textbf{Effectiveness of Patch Coherence Loss.}
The Patch Coherence Loss demonstrates its value by consistently enhancing model performance. It improves Top-1 accuracy by 1.2\%, 1.3\%, and 1.3\% for small, base, and large models, respectively. These improvements indicate that promoting coherence among patch representations strengthens the model's grasp of spatial relationships within tobacco product images, leading to more robust classification outcomes.

\subsection{Comparison with State-of-the-Art Methods}
\noindent\textbf{Tobacco Product Classification Task.}
We evaluated DEFEND on the PHAD dataset by fine-tuning only the linear classification layer. Our model outperformed ImageNet1K-pretrained architectures by up to 5.8\%, highlighting the importance of domain-specific pretraining, as shown in~\Cref{tab:phad_comparison}. It also surpassed recent self-supervised learning approaches pretrained on Tobacco-1M, demonstrating the effectiveness of our architecture and training strategy. Incorporating product descriptions further improved performance, achieving 78.3\% Top-1 and 83.1\% Top-5 accuracy. These results affirm the Tobacco Foundation Model's efficacy in generating rich, domain-specific representations for tobacco product classification tasks.

\begin{table}[h]
\centering
\caption{Classification results on PHAD dataset. Here, Desc.: Description.}
\label{tab:phad_comparison}
\small
\setlength{\tabcolsep}{4pt}
\begin{tabular}{lcccc}
\hline
\multirow{2}{*}{\textbf{Method}} & \multirow{2}{*}{\textbf{Desc.}} & \multirow{2}{*}{\begin{tabular}[c]{@{}c@{}}\textbf{Pre-train}\\\textbf{Data}\end{tabular}} & \multirow{2}{*}{\begin{tabular}[c]{@{}c@{}}\textbf{Acc@1}\\\textbf{(\%)}\end{tabular}} & \multirow{2}{*}{\begin{tabular}[c]{@{}c@{}}\textbf{Acc@5}\\\textbf{(\%)}\end{tabular}} \\
 & & & & \\
\hline
ResNet-50 \cite{he2016deep} & \xmark & ImageNet1K & 70.5 & 73.8 \\
EfficientNet \cite{tan2019efficientnet} & \xmark & ImageNet1K & 71.4 & 74.7 \\
ViT-B \cite{dosovitskiy2020image} & \xmark & ImageNet1K & 72.5 & 75.6 \\
\hline
DINO \cite{caron2021emerging} & \xmark & Tobacco-1M & 72.8 & 75.9 \\
MAE \cite{he2022masked} & \xmark & Tobacco-1M & 73.0 & 76.1 \\
CoCa \cite{yu2022coca} & \checkmark & Tobacco-1M & 73.2 & 76.4 \\
\hline
\textbf{DEFEND (Ours)} & \xmark & Tobacco-1M & 77.6 & 82.7 \\
\textbf{DEFEND (Ours)} & \checkmark & Tobacco-1M & \textbf{78.3} & \textbf{83.1} \\
\hline
\end{tabular}
\end{table}

\begin{figure}[h]
    \centering
    \includegraphics[width=0.9\linewidth]{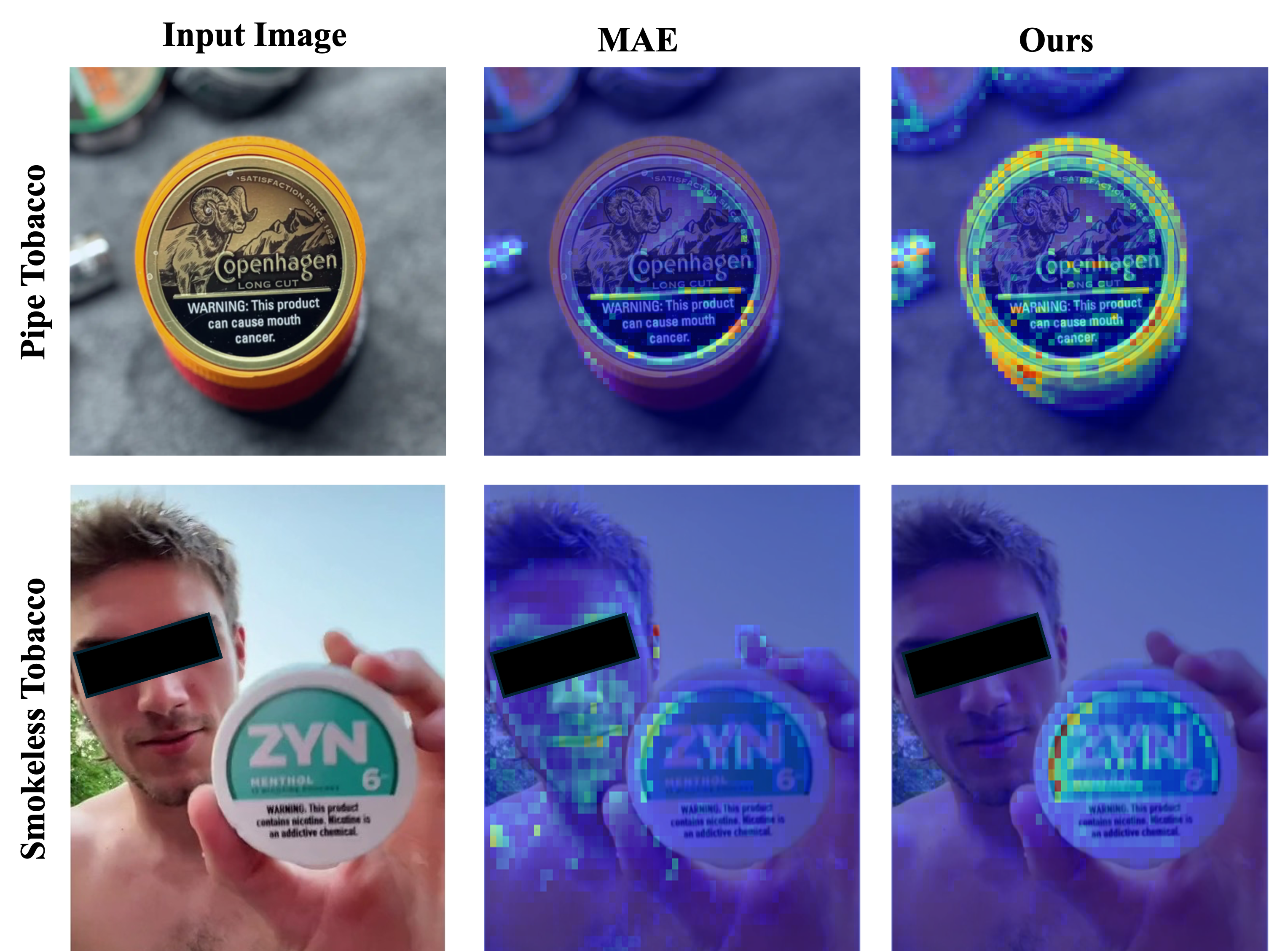}
    \caption{\textbf{Attention Visualization.} Compared to MAE~\cite{he2022masked}, our model demonstrates enhanced sensitivity to product-specific details in tobacco imagery. The model effectively highlights key product features and warning labels, maintaining robust attention even in challenging scenarios with varied backgrounds and user-generated content. \textbf{Best viewed in color.} 
    }
    \label{fig:vis1}
\end{figure}

\noindent\textbf{Visualization Results.}~\cref{fig:vis1} visualizes the attention maps of our model compared to MAE~\cite{he2022masked} pre-trained on the our proposed dataset. While MAE's attention disperses across background textures, DEFEND demonstrates precise focus on discriminative features such as warning labels, brand identifiers, and product contours. This targeted attention mechanism directly contributes to our model's superior classification performance, particularly in challenging scenarios with complex backgrounds or partial occlusions.

\noindent\textbf{Tobacco Products VQA Task.}
We evaluated DEFEND on the VQA task using the Tobacco-1M dataset, comparing it against state-of-the-art vision-language models across four question categories: Product Classification (PC), Usage Context (UC), Content Description (CD), and Health Impact (HI). All models were pre-trained on Tobacco-1M's train set and fine-tuned for the VQA task. Results on the test set are shown in Table \ref{tab:vqa_results}. For experimentation details, please refer to Sec. \textcolor{red}{E.1} of the Supp.
\begin{table}[h]
\centering
\caption{Visual Question Answering results on Tobacco-1M test set across question categories}
\label{tab:vqa_results}
\setlength{\tabcolsep}{4pt}
\small
\begin{tabular}{lcccccc}
\hline
& \multicolumn{5}{c}{\textbf{Accuracy (\%)}} & \textbf{F1} \\
\cline{2-6}
\textbf{Model} & \textbf{Overall} & \textbf{PC} & \textbf{UC} & \textbf{CD} & \textbf{HI} & \textbf{Score} \\
\hline
ViLBERT~\cite{lu2019vilbert} & 65.3 & 67.8 & 64.2 & 64.5 & 64.7 & 0.67 \\
CLIP~\cite{radford2021learning} & 66.9 & 69.2 & 65.8 & 65.9 & 66.7 & 0.68 \\
CoCa~\cite{yu2022coca} & 68.4 & 70.8 & 67.5 & 67.2 & 68.1 & 0.70 \\
Flamingo~\cite{alayrac2022flamingo} & 69.8 & 72.1 & 68.7 & 68.6 & 69.8 & 0.71 \\
MiniGPT-4~\cite{zhu2023minigpt} & 71.2 & 73.5 & 70.1 & 70.3 & 70.9 & 0.72 \\
MDETR~\cite{kamath2021mdetr} & 72.5 & 74.8 & 71.4 & 71.6 & 72.2 & 0.74 \\
\hline
\textbf{DEFEND (Ours)} & \textbf{73.8} & \textbf{76.2} & \textbf{72.9} & \textbf{72.8} & \textbf{73.3} & \textbf{0.75} \\
\hline
\end{tabular}
\small{\textbf{Note:} \emph{Overall} is the average of the given categories.}
\end{table}
Our model outperforms all baselines across question categories, with notable improvements in Product Classification (+1.4\%) and Usage Context (+1.5\%) compared to MDETR~\cite{kamath2021mdetr}. The consistent performance across categories demonstrates DEFEND's robust understanding of tobacco products' visual attributes, usage patterns, and health implications, making it particularly valuable for regulatory compliance and public health applications.

\noindent\textbf{Zero-shot Tobacco Product Classification.}
We tested our DEFEND's ability to classify novel tobacco products using the PHAD dataset as shown in~\Cref{tab:zero_shot}. The model leverages learned representations to match unseen product images with category descriptions. This approach allows for the identification of emerging tobacco products without additional training. Our model achieved 45.6\% accuracy, outperforming CLIP, CoCa, and MDETR. These results demonstrate the model's robust generalization capabilities and potential for adapting to evolving tobacco markets and regulatory challenges.
\begin{table}[h]
\centering
\caption{Zero-shot classification results on PHAD dataset.}
\label{tab:zero_shot}
\small
\begin{tabular}{lc}
\hline
\multirow{2}{*}{\textbf{Model}} & \multirow{2}{*}{\begin{tabular}[c]{@{}c@{}}\textbf{Accuracy}\\\textbf{(\%)}\end{tabular}} \\
&\\
\hline
CLIP \cite{radford2021learning} & 35.8 \\
CoCa \cite{yu2022coca} & 38.4 \\
MDETR \cite{kamath2021mdetr} & 40.9 \\
\textbf{DEFEND (Ours)} & \textbf{45.6} \\
\hline
\end{tabular}
\end{table}
%
\section{Conclusions}
This paper introduces Tobacco-1M, a comprehensive dataset of 1M tobacco product images with hierarchical annotations, and DEFEND, a novel foundation model for tobacco product understanding. By incorporating the Feature Enhancement Module, the Local-Global Visual Coherence, and the Enhanced Image-Text Alignment mechanisms, DEFEND achieves superior performance in product classification, marketing analysis, and health impact assessment. The model's strong zero-shot learning and visual question-answering capabilities demonstrate its effectiveness in identifying emerging tobacco products. 

\noindent\textbf{Limitations.} While DEFEND shows promising results, it has several limitations. The model's performance may be constrained when dealing with extremely novel tobacco products that significantly differ from those in the Tobacco-1M dataset. Its effectiveness across diverse cultural contexts and non-English packaging requires further investigation. In addition, the model's applicability to related public health domains without extensive retraining remains unexplored. Future work should address these limitations to enhance the model's versatility and real-world applicability in global tobacco control efforts.

{
    \small
    \bibliographystyle{ieeenat_fullname}
    \bibliography{main}
}

\end{document}


\maketitle

\section{Dataset Annotation Pipeline}

\subsection{Overview}
Our dataset annotation pipeline combines automated product categorization with expert manual curation by public health specialists to ensure comprehensive and accurate labeling of tobacco and nicotine products. The pipeline implements a multi-stage annotation process, incorporating domain expertise at critical validation points.

\subsection{Annotation Process}
The annotation process begins with initial data collection through the Apify platform~\cite{api}, which handles the preliminary query-based categorization and automated extraction of basic product information. During this phase, we perform initial filtering to ensure content relevance and basic quality standards.

Our automated categorization system then assigns preliminary labels for product categories, subcategories, brands, and basic attributes. This system also performs an initial classification of potential health impacts, creating a foundation for expert review.

\subsection{Expert Manual Curation}
The core of our annotation process relies on a diverse team of experts, including tobacco control specialists, public health researchers, healthcare professionals, and environmental impact assessors. These experts perform four key assessment areas:

\noindent\textbf{Health Impact Assessment:} Experts verify severity classifications, assess duration impacts, evaluate clinical relevance, and document specific health risks associated with each product.

\noindent\textbf{Usage Context Validation:} This involves verifying setting appropriateness, assessing regulatory compliance, evaluating social context, and documenting usage patterns.

\noindent\textbf{Marketing Analysis:} Experts review advertisement compliance, assess target audiences, evaluate promotional strategies, and verify regulatory adherence.

\noindent\textbf{Environmental Impact Evaluation:} This includes waste categorization, environmental hazard assessment, long-term impact analysis, and sustainability evaluation.

\subsection{Quality Control Framework}
Our quality control framework centers on three primary components: inter-annotator agreement, validation protocols, and documentation standards. We maintain high data quality through multiple expert reviews per entry and consensus-based resolution of disagreements. Regular calibration sessions and standardized annotation guidelines ensure consistency across the dataset.

\subsection{Technical Infrastructure}
The annotation process is supported by a custom-built web-based platform that enables real-time collaboration and includes integrated validation tools. Our quality assurance system features automated consistency checks, duplicate detection, and error flagging capabilities. All data is managed through a centralized database with robust backup protocols and comprehensive access controls.

\subsection{Continuous Improvement}
The annotation pipeline maintains its effectiveness through regular refinement of processes and standards. We continuously review and update annotation guidelines, classification criteria, and quality control measures based on expert feedback and emerging research. This iterative improvement ensures the dataset remains current and valuable for both research and public health applications.

Through this comprehensive approach, our annotation pipeline creates a high-quality, expertly curated dataset that serves both research and public health purposes while maintaining rigorous standards for accuracy and relevance.

\section{Dataset Structure and Organization}
\subsection{Data Format}
Our tobacco product dataset follows a structured annotation format designed to capture comprehensive information about tobacco and nicotine products, their health impacts, usage contexts, and environmental implications. The basic structure is organized as follows:

\begin{lstlisting}[language=json,escapechar=!,backgroundcolor=\color{white}, mathescape=true]
data[{
    "query": str,
    "imageUrl": str,
    "image_id": str,
    "category": str,
    "sub-category": str,
    "health_impact_labels": {
        "severity": [{
            "level": str,
            "impact": str,
            "visual_cues": [str]
        }],
        "duration": [{
            "type": str,
            "effects": [str],
            "visual_indicators": [str]
        }]
    },
    "usage_context": {
        "settings": [{
            "type": str,
            "visual_cues": [str]
        }],
        "regulatory_zones": [{
            "type": str,
            "visual_cues": [str]
        }]
    },
    "content_purpose": {
        "marketing": [{
            "type": str,
            "visual_elements": [str]
        }],
        "regulatory": [{
            "type": str,
            "visual_elements": [str]
        }]
    },
    "environmental_impact": {
        "type": str,
        "description": str,
        "visual_indicators": {
            "litter": [str],
            "pollution": [str]
        }
    }
}]
\end{lstlisting}

\subsection{Basic Information}
Each entry in the dataset contains fundamental attributes:
\begin{itemize}
    \item \texttt{query}: Product description or search term used for identification
    \item \texttt{imageUrl}: URL reference to the product image
    \item \texttt{image\_id}: Unique identifier for the image
    \item \texttt{category}: Main product classification
    \item \texttt{sub-category}: Specific product type within the main category
\end{itemize}

\subsection{Product Categories}
The dataset encompasses three primary product categories:
\begin{enumerate}
    \item \textbf{Combustible Tobacco Products}
    \begin{itemize}
        \item Cigarettes
        \item Cigars
        \item Pipe Tobacco
        \item Hookah/Shisha
    \end{itemize}
    
    \item \textbf{Non-Combustible Products}
    \begin{itemize}
        \item E-cigarettes/Vapes
        \item Smokeless Tobacco
        \item Heated Tobacco
    \end{itemize}
    
    \item \textbf{Nicotine Replacement Therapy}
    \begin{itemize}
        \item Patches
        \item Gums
        \item Lozenges
    \end{itemize}
\end{enumerate}

\subsection{Health Impact Labels}
The health impact annotations are structured to capture comprehensive information about both immediate and long-term health consequences:

\subsubsection{Severity Classification}
\begin{itemize}
    \item \textbf{High Severity}
    \begin{itemize}
        \item \textit{Lung Cancer}: Primary association with combustible products
        \begin{itemize}
            \item Documented in cigarettes, cigars, pipe tobacco
            \item Progressive respiratory system damage
            \item Associated mortality rates
        \end{itemize}
        \item \textit{Cardiovascular Disease}: Across all product categories
        \begin{itemize}
            \item Heart disease manifestations
            \item Stroke risk factors
            \item Blood pressure implications
        \end{itemize}
        \item \textit{Oral Cancer}: Prevalent in specific products
        \begin{itemize}
            \item Smokeless tobacco correlations
            \item Heated tobacco implications
            \item Mouth and throat cancer risks
        \end{itemize}
    \end{itemize}
    
    \item \textbf{Medium Severity}
    \begin{itemize}
        \item \textit{Respiratory Issues}
        \begin{itemize}
            \item E-cigarette/vaping-associated lung injury (EVALI)
            \item Chronic bronchitis
            \item Reduced lung function
        \end{itemize}
        \item \textit{Oral Health Problems}
        \begin{itemize}
            \item Gum disease progression
            \item Tooth decay patterns
            \item Mouth lesions
        \end{itemize}
    \end{itemize}
    
    \item \textbf{Low Severity}
    \begin{itemize}
        \item \textit{Temporary Effects}
        \begin{itemize}
            \item Skin irritation from patches
            \item Mouth irritation from gums/lozenges
            \item Minor respiratory irritation
        \end{itemize}
    \end{itemize}
\end{itemize}

\subsubsection{Duration Categories}
\begin{itemize}
    \item \textbf{Short-term Effects}
    \begin{itemize}
        \item \textit{Immediate Physiological Responses}
        \begin{itemize}
            \item Increased heart rate
            \item Blood pressure elevation
            \item Respiratory irritation
        \end{itemize}
        \item \textit{Acute Symptoms}
        \begin{itemize}
            \item Nausea and dizziness
            \item Headaches
            \item Sleep disturbances
        \end{itemize}
    \end{itemize}
    
    \item \textbf{Long-term Effects}
    \begin{itemize}
        \item \textit{Chronic Conditions}
        \begin{itemize}
            \item Cancer development patterns
            \item Cardiovascular disease progression
            \item Chronic respiratory conditions
        \end{itemize}
        \item \textit{Systemic Impact}
        \begin{itemize}
            \item Immune system deterioration
            \item Organ system damage
            \item Addiction patterns
        \end{itemize}
    \end{itemize}
\end{itemize}

\subsection{Usage Context}
Usage context labels capture the environmental and social dimensions of tobacco product use:

\subsubsection{Setting Classifications}
\begin{itemize}
    \item \textbf{Individual Use}
    \begin{itemize}
        \item \textit{Private Spaces}
        \begin{itemize}
            \item Personal residences
            \item Private vehicles
            \item Individual offices
        \end{itemize}
        \item \textit{Usage Patterns}
        \begin{itemize}
            \item Solitary consumption
            \item Personal rituals
            \item Frequency patterns
        \end{itemize}
    \end{itemize}
    
    \item \textbf{Social Settings}
    \begin{itemize}
        \item \textit{Group Environments}
        \begin{itemize}
            \item Social gatherings
            \item Recreational venues
            \item Community spaces
        \end{itemize}
        \item \textit{Behavioral Patterns}
        \begin{itemize}
            \item Social interactions
            \item Group consumption patterns
            \item Peer influence factors
        \end{itemize}
    \end{itemize}
    
    \item \textbf{Public Spaces}
    \begin{itemize}
        \item \textit{Open Areas}
        \begin{itemize}
            \item Parks and recreational spaces
            \item Street environments
            \item Commercial zones
        \end{itemize}
        \item \textit{Regulatory Considerations}
        \begin{itemize}
            \item Distance requirements
            \item Designated areas
            \item Public health guidelines
        \end{itemize}
    \end{itemize}
\end{itemize}

\subsection{Content Purpose}
The content purpose classification system encompasses both marketing and regulatory aspects:

\subsubsection{Marketing Elements}
\begin{itemize}
    \item \textbf{Direct Advertisement}
    \begin{itemize}
        \item \textit{Product Presentation}
        \begin{itemize}
            \item Brand prominence
            \item Product features
            \item Marketing messages
        \end{itemize}
        \item \textit{Target Audience Indicators}
        \begin{itemize}
            \item Demographic focus
            \item Lifestyle associations
            \item Cultural references
        \end{itemize}
    \end{itemize}
    
    \item \textbf{Product Placement}
    \begin{itemize}
        \item \textit{Contextual Integration}
        \begin{itemize}
            \item Natural use scenarios
            \item Lifestyle contexts
            \item Environmental setting
        \end{itemize}
        \item \textit{Promotional Strategy}
        \begin{itemize}
            \item Subtle brand presence
            \item Associated activities
            \item Social context implications
        \end{itemize}
    \end{itemize}
\end{itemize}

\subsubsection{Regulatory Elements}
\begin{itemize}
    \item \textbf{Warning Labels}
    \begin{itemize}
        \item \textit{Health Warnings}
        \begin{itemize}
            \item Mandatory text content
            \item Visual warning elements
            \item Size and placement requirements
        \end{itemize}
        \item \textit{Safety Information}
        \begin{itemize}
            \item Usage guidelines
            \item Risk statements
            \item Emergency information
        \end{itemize}
    \end{itemize}
    
    \item \textbf{Compliance Indicators}
    \begin{itemize}
        \item \textit{Age Restrictions}
        \begin{itemize}
            \item Minimum age requirements
            \item Purchase restrictions
            \item Verification systems
        \end{itemize}
        \item \textit{Usage Regulations}
        \begin{itemize}
            \item Legal requirements
            \item Public space restrictions
            \item Sale limitations
        \end{itemize}
    \end{itemize}
\end{itemize}

\subsection{Environmental Impact}
Environmental impact labels document the ecological consequences of tobacco products:

\subsubsection{Combustible Products}
\begin{itemize}
    \item \textbf{Direct Environmental Effects}
    \begin{itemize}
        \item \textit{Air Pollution}
        \begin{itemize}
            \item Smoke emissions
            \item Atmospheric pollutants
            \item Indoor air quality
        \end{itemize}
        \item \textit{Solid Waste}
        \begin{itemize}
            \item Cigarette butts
            \item Packaging materials
            \item Associated debris
        \end{itemize}
    \end{itemize}
    
    \item \textbf{Long-term Environmental Impact}
    \begin{itemize}
        \item \textit{Ecosystem Effects}
        \begin{itemize}
            \item Soil contamination
            \item Water system pollution
            \item Wildlife impact
        \end{itemize}
        \item \textit{Degradation Patterns}
        \begin{itemize}
            \item Decomposition timelines
            \item Chemical leaching
            \item Accumulation patterns
        \end{itemize}
    \end{itemize}
\end{itemize}

\subsubsection{Non-Combustible Products}
\begin{itemize}
    \item \textbf{Electronic Waste}
    \begin{itemize}
        \item \textit{Device Components}
        \begin{itemize}
            \item Battery disposal
            \item Electronic parts
            \item Circuit boards
        \end{itemize}
        \item \textit{Consumable Elements}
        \begin{itemize}
            \item Cartridge disposal
            \item Liquid containers
            \item Packaging waste
        \end{itemize}
    \end{itemize}
    
    \item \textbf{Chemical Waste}
    \begin{itemize}
        \item \textit{Liquid Waste}
        \begin{itemize}
            \item Nicotine solutions
            \item Flavoring compounds
            \item Chemical residues
        \end{itemize}
        \item \textit{Environmental Persistence}
        \begin{itemize}
            \item Biodegradability factors
            \item Toxic accumulation
            \item Groundwater impact
        \end{itemize}
    \end{itemize}
\end{itemize}


This comprehensive labeling system enables detailed analysis of tobacco products, their health implications, usage patterns, and environmental impacts, supporting both public health research and regulatory compliance assessment.

\subsection{Python Script to Download and Organize the Dataset}
Please use the \texttt{download.py} script file (provided in the .zip file of supplementary material) to download all the images into their corresponding \texttt{'category'} directories. The dataset json file is also provided in the .zip file of supplementary material, also please note that \emph{we provide only a subset of our whole dataset due to the cap of 200MB for the supplementary material}. However, the whole dataset will be available as a single json file which is 920MB and after downloading the whole dataset, the total (1M) images will be around 350GB.

\section{Inference}
During inference, we only utilize the teacher model following the common practice in EMA-based knowledge distillation approaches~\cite{yu2022coca, caron2021emerging}. This decision is based on the teacher's more stable and robust feature representations achieved through EMA updates during training.

Given a test image $I$, the inference process proceeds as follows:
\begin{align}
    \bm{f_G} &= E_{global}(I) \\
    \hat{f}_G &= A_G(\bm{f_G}, \bm{f_G}, \bm{f_G}) \\
    \bm{f_G^*} &= F_G(A_{VT}(\hat{f}_G, \hat{f}_T, \hat{f}_T))
\end{align}
where the enhanced global features $\bm{f_G^*}$ are used for downstream tasks such as classification or visual question answering. For tasks requiring text inputs (e.g., VQA), we similarly process the text through the text encoder and feature enhancement module:
\begin{align}
    \bm{f_T} &= E_{text}(T) \\
    \hat{f}_T &= A_T(\bm{f_T}, \bm{f_T}, \bm{f_T}) \\
    \bm{f_T^*} &= F_T(A_{TV}(\hat{f}_T, \hat{f}_G, \hat{f}_G))
\end{align}

This streamlined inference process maintains model efficiency while leveraging the robust representations learned through our teacher-student framework. The student network, having served its purpose in training by providing local feature guidance, is not needed during inference.

\section{Implementation Details}

\subsection{Adaptive Sampling Strategy}
Our adaptive sampling strategy $\psi$ selects informative patches based on their saliency scores. For each patch p\_i, we compute a saliency score s\_i as:

\begin{equation}
    s_i = \alpha \cdot \text{EdgeScore}(p_i) + \beta \cdot \text{TextScore}(p_i) + \gamma \cdot \text{ContrastScore}(p_i)
\end{equation}

where:
- EdgeScore: Measures structural information using Sobel edge detection
- TextScore: Evaluates text presence using EAST text detection
- ContrastScore: Calculates local contrast with surrounding patches
- $\alpha, \beta, \gamma$ are weighting parameters set to 0.4, 0.4, and 0.2 respectively

Patches are selected if their saliency score exceeds the threshold $\lambda$ = 0.3. This typically results in retaining 70-80\% of patches, with higher retention rates for patches containing text, logos, and warning labels.

\subsection{Hyperparameter Selection}
We performed grid search over the following hyperparameter ranges which is presented in~\cref{tab:hyperparams}.

\begin{table*}[h]
\centering
\caption{Hyperparameter search space and final values}
\label{tab:hyperparams}
\setlength{\tabcolsep}{4pt}
\begin{tabular}{lccc}
\hline
\textbf{Hyperparameter} & \textbf{Search Range} & \textbf{Final Value} & \textbf{Selection Criteria} \\
\hline
Learning Rate & [1e-4, 5e-4, 1e-3] & 1e-3 & Convergence speed \\
Batch Size & [64, 128, 256] & 128 & Memory efficiency \\
Temperature $\tau$ & [0.05, 0.07, 0.1] & 0.07 & Contrastive learning stability \\
Dropout Rate & [0.1, 0.2, 0.3] & 0.2 & Validation accuracy \\
Weight Decay & [0.01, 0.05, 0.1] & 0.05 & Regularization effect \\
\hline
\end{tabular}
\end{table*}

\subsection{Training Procedure Details}
Our training procedure consists of three phases:

\noindent\textbf{Phase 1: Warm-up (20 epochs)}
\begin{itemize}
    \item Linear learning rate warm-up from 0 to 1e-3
    \item Only global feature encoder trained
    \item Basic augmentations applied
\end{itemize}
\noindent\textbf{Phase 2: Main Training (200 epochs)}
\begin{itemize}

\item Cosine learning rate schedule with base lr=1e-3
\item All components trained jointly
\item Full augmentation strategy:
\begin{itemize}
    \item Random horizontal flip (p=0.5)
  \item Random rotation (±10°)
  \item Color jittering (brightness=0.2, contrast=0.2)
  \item  Random cropping (0.85-1.0 of original size)
  \item  Gaussian blur (p=0.3)
  \end{itemize}
\end{itemize}
\noindent\textbf{Phase 3: Fine-tuning (80 epochs)}
\begin{itemize}
    \item Reduced learning rate (1e-4)
\item Focus on alignment losses
\item  Gradient accumulation every 4 steps
\end{itemize}
\subsection{Data Augmentation Details}
Our augmentation strategy is carefully designed for tobacco product characteristics:

\begin{itemize}
\item \textbf{Geometric Transformations:}
    \begin{itemize}
        \item Scale: [0.85, 1.0] of original size
    \item Rotation: ±10° (limited to preserve text readability)
    \item  Translation: ±10\% of image size
     \end{itemize}
\item \textbf{Appearance Transformations:}
    \begin{itemize}
        \item Brightness: [0.8, 1.2]
    \item  Contrast: [0.8, 1.2]
    \item  Saturation: [0.8, 1.2]
    \item  Hue: [-0.1, 0.1]
    \end{itemize}
\item \textbf{Noise and Filtering:}
    \begin{itemize}
    \item Gaussian noise ($\sigma$ = 0.01)
    \item Gaussian blur (kernel size = 3, $\sigma$ = 0.5)
    \item JPEG compression quality: [70, 100]
    \end{itemize}
\end{itemize}

\subsection{Model Initialization}
\begin{itemize}
    \item Visual backbone: Initialized from ImageNet-pretrained weights
\item  Text encoder: Initialized from BERT-base

\end{itemize}
\subsection{Hardware and Software Details}
\begin{itemize}
    \item Training Infrastructure:
  \begin{itemize}
      \item 4 NVIDIA A100 GPUs (40GB each)
  \item 2 AMD EPYC 7763 64-Core Processors
  \item 512GB RAM
  \end{itemize}
  
\item  Software Environment:
  \begin{itemize}
      \item PyTorch 1.12.0
  \item  CUDA 11.6
  \item  Python 3.8
  \item  Torchvision 0.13.0
 \end{itemize}
 \end{itemize}

\section{More Ablation Studies}

We performed additional ablation studies as observed in~\cref{tab:arch_ablation} and their analysis is as follows,
\begin{table}[h]
\centering
\caption{Impact of different architectural choices on model performance. PE: Positional Encoding, TL: Transformer Layers}
\label{tab:arch_ablation}
\setlength{\tabcolsep}{4pt}
\begin{tabular}{ccccc}
\hline
\textbf{Configuration} & \textbf{Acc@1} & \textbf{Acc@5} & \begin{tabular}[c]{@{}c@{}}\textbf{Training}\\\textbf{Time (h)}\end{tabular} & \begin{tabular}[c]{@{}c@{}}\textbf{Params}\\\textbf{(M)}\end{tabular} \\
\hline
&\multicolumn{3}{c}{\textit{Patch Size}}& \\
\hline
8 × 8 & 77.4 & 82.3 & 36 & 102 \\
16 × 16 & \textbf{78.3} & \textbf{83.1} & 24 & 86 \\
32 × 32 & 76.8 & 81.5 & 18 & 75 \\
\hline
&\multicolumn{3}{c}{\textit{Transformer Layers}}& \\
\hline
6 layers & 76.2 & 81.1 & 16 & 68 \\
12 layers & \textbf{78.3} & \textbf{83.1} & 24 & 86 \\
18 layers & 78.1 & 82.9 & 38 & 112 \\
\hline
&\multicolumn{3}{c}{\textit{Positional Encoding}}& \\
\hline
Absolute & 77.6 & 82.4 & 24 & 86 \\
Relative & \textbf{78.3} & \textbf{83.1} & 24 & 86 \\
Learnable & 77.9 & 82.7 & 24 & 87 \\
\hline
\end{tabular}
\end{table}

\subsection{Analysis of Architectural Choices}

\noindent\textbf{Impact of Patch Size.} We experiment with different patch sizes to balance feature granularity and computational efficiency. While 8×8 patches capture finer details, they significantly increase computational cost and memory usage. The 16×16 patch size achieves the best trade-off, maintaining high accuracy while reducing training time by 33\% compared to 8×8 patches. Larger 32×32 patches show degraded performance (-1.5\% in Acc@1), likely due to loss of fine-grained features crucial for distinguishing similar tobacco products.

\noindent\textbf{Transformer Layer Depth.} We investigate the impact of model depth on performance. While deeper networks (18 layers) show comparable accuracy (-0.2\%), they increase training time by 58\% and parameter count by 30\%. The 12-layer configuration provides the optimal balance between performance and efficiency. Models with 6 layers show notable performance degradation (-2.1\%), indicating insufficient capacity for complex tobacco product understanding.

\noindent\textbf{Positional Encoding Schemes.} We compare three positional encoding strategies: absolute, relative, and learnable positions. Relative positional encoding demonstrates superior performance, suggesting the importance of spatial relationships in tobacco product recognition. Learnable positions show competitive results (-0.4\%) but require additional parameters without significant performance gains. Absolute positions perform adequately but lag behind in accuracy by 0.7\%, particularly for larger products that span multiple patches.

\section{Comprehensive Performance Analysis}

\begin{table}[h]
\centering
\caption{Performance breakdown across different product categories and their variants}
\label{tab:category_analysis}
\setlength{\tabcolsep}{4pt}
\begin{tabular}{lccc}
\hline
\textbf{Product Category} & \begin{tabular}[c]{@{}c@{}}\textbf{Sample}\\\textbf{Count}\end{tabular} & \begin{tabular}[c]{@{}c@{}}\textbf{Acc}\\\textbf{(\%)}\end{tabular} & \begin{tabular}[c]{@{}c@{}}\textbf{Error}\\\textbf{Types}\end{tabular} \\
\hline
\multicolumn{4}{c}{\textit{Combustible Products}} \\
Cigarettes & 300K & 88.4 & Brand confusion \\
Cigars & 100K & 86.2 & Size variation \\
Hookah & 50K & 82.1 & Material similarity \\
\hline
\multicolumn{4}{c}{\textit{Non-Combustible Products}} \\
E-cigarettes & 120K & 85.7 & Device types \\
Smokeless & 50K & 83.9 & Packaging style \\
Heated tobacco & 30K & 81.5 & Novel designs \\
\hline
\multicolumn{4}{c}{\textit{Nicotine Replacement}} \\
Patches & 25K & 89.2 & Clear packaging \\
Gum & 25K & 87.8 & Brand variants \\
\hline
\end{tabular}
\end{table}

\begin{table}[h]
\centering
\caption{Computational efficiency comparison with baseline methods}
\label{tab:computation_analysis}
\setlength{\tabcolsep}{4pt}
\begin{tabular}{lcccc}
\hline
\textbf{Model} & \begin{tabular}[c]{@{}c@{}}\textbf{Params}\\\textbf{(M)}\end{tabular} & \begin{tabular}[c]{@{}c@{}}\textbf{FLOPs}\\\textbf{(G)}\end{tabular} & \begin{tabular}[c]{@{}c@{}}\textbf{Inference}\\\textbf{Time (ms)}\end{tabular} & \begin{tabular}[c]{@{}c@{}}\textbf{GPU}\\\textbf{Mem (GB)}\end{tabular} \\
\hline
CLIP~\cite{radford2021learning} & 68 & 12.5 & 18.3 & 4.2 \\
CoCa~\cite{yu2022coca} & 75 & 14.2 & 22.1 & 5.1 \\
MDETR~\cite{kamath2021mdetr} & 82 & 15.8 & 25.4 & 5.8 \\
\textbf{DEFEND} & 86 & 16.3 & 24.7 & 4.3 \\
\hline
\end{tabular}
\end{table}




\noindent\textbf{Category-wise Performance.} Table~\ref{tab:category_analysis} presents a detailed breakdown of our model's performance across different product categories. Notably, traditional products like cigarettes and nicotine replacement products show higher accuracy (88.4\% and 89.2\% respectively), likely due to more standardized packaging and clearer visual features. E-cigarettes and novel tobacco products present greater challenges, with accuracy dropping to 81.5-85.7\%. This performance gap primarily stems from rapid product innovation and diverse design variations in these categories. Interestingly, the most common error types vary significantly: traditional cigarettes mainly suffer from brand confusion due to similar packaging designs, while e-cigarettes show confusion between different device types, particularly with newer form factors.

\noindent\textbf{Computational Efficiency.} As shown in Table~\ref{tab:computation_analysis}, DEFEND achieves competitive efficiency despite its enhanced feature processing capabilities. While our model has marginally higher computational requirements compared to baselines (16.3G FLOPs vs. 12.5-15.8G), the increase is modest (+3.2\% in inference time compared to MDETR) given the significant performance gains. The memory footprint remains practical for deployment, requiring 4.3 GB GPU memory during inference, making it suitable for standard GPU hardware. This efficiency-performance trade-off is particularly important for real-time tobacco product monitoring applications.

\noindent\textbf{Statistical Significance.} To validate our improvements over baseline methods, we conducted paired t-tests across multiple runs. The improvements in classification accuracy are statistically significant (p < 0.01) across all major product categories. The performance gains are most consistent in traditional product categories ($sigma$ = 0.3\%) and show slightly higher variance in emerging product categories ($sigma$ = 0.7\%), reflecting the inherent challenges in novel product recognition.

\subsection{Visual Question Answering Task Setup}

\noindent\textbf{Question Categories.} We evaluate our model on four distinct types of questions:
\begin{itemize}
    \item \textbf{Product Classification (PC):} Questions about product type, brand, and variants (e.g., "What type of e-cigarette is shown?", "Which brand of cigarettes is this?")
    \item \textbf{Usage Context (UC):} Questions about product usage and consumption methods (e.g., "How is this product typically used?", "What is the intended method of consumption?")
    \item \textbf{Content Description (CD):} Questions about packaging elements and warnings (e.g., "What health warnings are visible?", "What nicotine content is listed?")
    \item \textbf{Health Impact (HI):} Questions about health risks and safety information (e.g., "What specific health risks are mentioned?", "Are there any addiction warnings?")
\end{itemize}

\noindent\textbf{Dataset Construction.} For VQA evaluation, we created a test set of 10,000 question-answer pairs:
\begin{itemize}
    \item 2,500 questions per category (PC, UC, CD, HI)
    \item Questions manually curated by tobacco control experts
    \item Each question has a ground truth answer and acceptable alternative answers
    \item Questions span across all major product categories in Tobacco-1M
\end{itemize}

\begin{table}[h]
\centering
\caption{Distribution of question types and answer formats in VQA evaluation}
\label{tab:vqa_distribution}
\setlength{\tabcolsep}{3pt}
\small
\begin{tabular}{lcccc}
\hline
\textbf{Question} & \textbf{Short} & \textbf{Yes/No} & \textbf{Multiple} & \textbf{Total} \\
\textbf{Category} & \textbf{Answer} & \textbf{Answer} & \textbf{Choice} & \textbf{Questions} \\
\hline
Product Classification & 1,500 & 500 & 500 & 2,500 \\
Usage Context & 1,200 & 800 & 500 & 2,500 \\
Content Description & 1,800 & 200 & 500 & 2,500 \\
Health Impact & 1,400 & 600 & 500 & 2,500 \\
\hline
\textbf{Total} & 5,900 & 2,100 & 2,000 & 10,000 \\
\hline
\end{tabular}
\end{table}

\noindent\textbf{Evaluation Metrics.}
We employ multiple metrics to comprehensively evaluate VQA performance:
\begin{itemize}
    \item \textbf{Accuracy:} Exact match between predicted and ground truth answers
    \item \textbf{F1 Score:} For assessing precision and recall of answer content
\end{itemize}

\noindent\textbf{Implementation Details.}
The VQA component is implemented as follows:
\begin{itemize}
    \item Question encoder: BERT-base with max length of 64 tokens
    \item Answer decoder: Transformer decoder with 6 layers
    \item Cross-attention between visual and textual features
    \item Beam search decoding with beam size 4
    \item Maximum answer length: 30 tokens
\end{itemize}


\noindent\textbf{Qualitative Analysis.}
Table~\ref{tab:vqa_examples} shows representative examples of model responses:

\begin{table*}[h]
\centering
\caption{Example VQA responses across different question categories}
\label{tab:vqa_examples}
\small
\begin{tabular}{llll}
\hline
\textbf{Question Type} & \textbf{Question} & \textbf{Model Response} & \textbf{Analysis} \\
\hline
Product Classification & "What brand and type of  & "JUUL pod device in & Correct identification of \\
& e-cigarette is this?" & mint flavor" & brand and variant \\
\hline
Usage Context & "How should this product & "Insert pod into device, & Detailed understanding of \\
& be used?" & inhale through mouthpiece" & product usage \\
\hline
Content Description & "What warnings are & "Contains nicotine, & Accurate extraction of \\
& visible on the package?" & highly addictive substance" & warning information \\
\hline
Health Impact & "What health risks are & "Risk of nicotine addiction & Comprehensive health \\
& mentioned?" & and lung damage" & risk identification \\
\hline
\end{tabular}
\end{table*}

\section{Ethical Considerations and Potential Misuse}

While our proposed framework is developed to support public health surveillance and tobacco control efforts, we acknowledge several important ethical considerations and potential risks of misuse that warrant careful discussion.

\subsection{Privacy and Data Protection}
The development of Tobacco-1M adhered to strict privacy protection protocols:
\begin{itemize}
    \item All images were collected from public sources, with personally identifiable information carefully removed through automated and manual verification processes
    \item Rigorous filtering mechanisms were implemented to exclude images containing minors or vulnerable populations
    \item The dataset focuses exclusively on product images, deliberately avoiding sensitive user-generated content
\end{itemize}

\subsection{Potential Misuse Scenarios}
We identify two primary categories of potential misuse:

\noindent\textbf{Commercial Exploitation:}
The framework's capabilities could potentially be repurposed for marketing optimization:
\begin{itemize}
    \item Product feature analysis could reveal consumer preferences
    \item Zero-shot learning capabilities might be used to test novel marketing approaches
    \item Advanced visual understanding could inform product design strategies
\end{itemize}

\noindent\textbf{Regulatory Circumvention:}
The system's detection mechanisms could be studied to design evasive packaging:
\begin{itemize}
    \item Knowledge of detection patterns might enable circumvention attempts
    \item Model interpretability could reveal detection blind spots
    \item Transfer learning capabilities might be exploited for adversarial purposes
\end{itemize}

\subsection{Mitigation Strategies}
To address these concerns, we implement multiple safeguards:
\begin{itemize}
    \item Dataset access is restricted to verified research institutions and public health organizations
    \item Usage agreements explicitly prohibit applications promoting tobacco products
    \item Regular auditing protocols monitor model applications and outcomes
    \item Continuous assessment of potential misuse patterns
    \item Integration of feedback from tobacco control experts
\end{itemize}

\subsection{Public Health Impact}
The framework is designed as a complementary tool for existing tobacco control efforts:
\begin{itemize}
    \item Technology augments, rather than replaces, human expertise
    \item Regular impact assessments ensure alignment with public health objectives
    \item System limitations are transparently documented and communicated
\end{itemize}

We acknowledge that these considerations may not be exhaustive and welcome community feedback on additional ethical implications and mitigation strategies. Future work will continue to refine these safeguards as new challenges emerge.

 {\small
 \bibliographystyle{ieeenat_fullname}
 \bibliography{main}
 }